\documentclass[conference]{IEEEtran}
\IEEEoverridecommandlockouts
\usepackage{cite}
\usepackage{amsmath,amssymb,amsfonts}
\usepackage{algorithmic}
\usepackage{bbding}
\usepackage{graphicx}
\usepackage{booktabs}
\usepackage{multirow}
\usepackage{textcomp}
\usepackage{xcolor}
\usepackage{float}
\usepackage{subfigure}
\usepackage{makecell}
\let\OLDthebibliography\thebibliography
\renewcommand\thebibliography[1]{
  \OLDthebibliography{#1}
  \setlength{\parskip}{0pt}
  \setlength{\itemsep}{0pt plus 0.3ex}
}

\def\BibTeX{{\rm B\kern-.05em{\sc i\kern-.025em b}\kern-.08em
    T\kern-.1667em\lower.7ex\hbox{E}\kern-.125emX}}

\newcommand{\linebreakand}{%
\end{@IEEEauthorhalign}
\hfill\mbox{}\par
\mbox{}\hfill\begin{@IEEEauthorhalign}
}
\makeatother

\begin{document}

\title{Transmission and Color-guided Network for Underwater Image Enhancement\\
 \thanks{
This work is supported by Natural Science Foundation of China (Grant No. U20A20196, 62202429) and Zhejiang Provincial Natural Science Foundation of China under Grant No. LR21F020002, LY23F020024.}
\thanks{$*$ The corresponding author.}
}

\author{\IEEEauthorblockN{Pan Mu}
\IEEEauthorblockA{\textit{College of Computer Science \& Technology} \\
\textit{Zhejiang University of Technology}\\
Hangzhou, China \\
panmu@zjut.edu.cn}
\and
\IEEEauthorblockN{Jing Fang}
\IEEEauthorblockA{\textit{College of Computer Science \& Technology} \\
\textit{Zhejiang University of Technology}\\
Hangzhou, China \\
211122120024@zjut.edu.cn}
\and
\quad\quad\quad\quad\quad\quad
\IEEEauthorblockN{Haotian Qian}
\IEEEauthorblockA{\quad\quad\quad\quad \quad\quad
	\textit{College of Computer Science \& Technology} \\
	\quad\quad\quad\quad\quad\quad
\textit{Zhejiang University of Technology}\\
\quad\quad\quad\quad\quad\quad
Hangzhou, China \\
\quad\quad\quad\quad\quad\quad
201906062215@zjut.edu.cn}
\and
\quad\quad\quad\quad\quad
\IEEEauthorblockN{Cong Bai*}
\IEEEauthorblockA{\quad\quad\quad\quad\quad
	\textit{College of Computer Science \& Technology} \\
\quad\quad\quad\quad\quad
\textit{Zhejiang University of Technology}\\
\quad\quad\quad\quad\quad
Hangzhou, China \\
\quad\quad\quad\quad\quad
congbai@zjut.edu.cn}
}

\maketitle

\begin{abstract}
In recent years, with the continuous development of the marine industry, underwater image enhancement has attracted plenty of attention. Unfortunately, the propagation of light in water will be absorbed by water bodies and scattered by suspended particles, resulting in color deviation and low contrast. To solve these two problems, we propose an Adaptive Transmission and Dynamic Color guided network (named ATDCnet) for underwater image enhancement. In particular, to exploit the knowledge of physics, we design an Adaptive Transmission-directed Module (ATM) to better guide the network. To deal with the color deviation problem, we design a Dynamic Color-guided Module (DCM) to post-process the enhanced image color. Further, we design an Encoder-Decoder-based Compensation (EDC) structure with attention and a multi-stage feature fusion mechanism to perform color restoration and contrast enhancement simultaneously. Extensive experiments demonstrate the state-of-the-art performance of the ATDCnet on multiple benchmark datasets. 
\end{abstract}


\begin{figure*}[t]
	\centering
	\begin{center}
		\begin{tabular}{c@{\extracolsep{0.2mm}}}
			\includegraphics[width=1\linewidth]{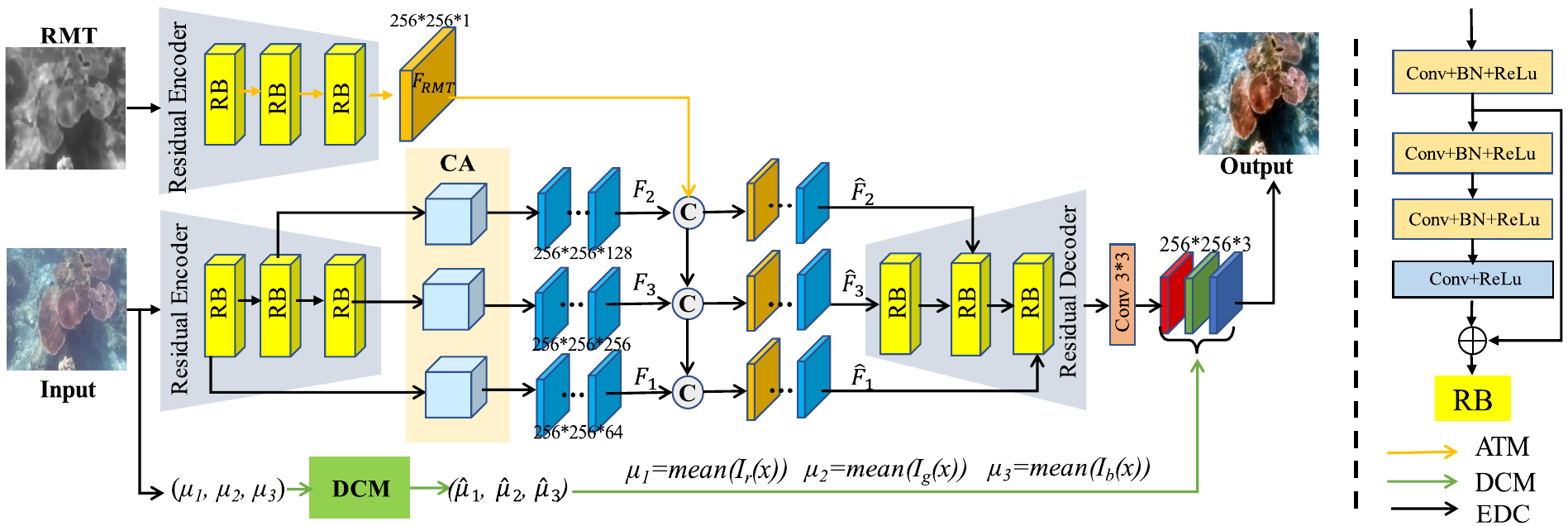}
		\end{tabular}
	\end{center}
	\vspace{-5mm}
	\caption{The overall framework of ATDCnet which composed of three branches. The Residual Block (i.e., RB) is mainly composed of some convolution layer, BatchNorm layer, and LeakyReLU layer. The reverse medium transmission map (denoted as RMT) represents the transmission information of the underwater image. $(\mu_1,\mu_2,\mu_3)$ characterize the general color information of the underwater image.}
	\label{fig:1}
\end{figure*}

\section{Introduction} 
Underwater images play an important role in the marine industry, such as underwater archaeology~\cite{bailey2008archaeology} and underwater target detection~\cite{dubreuil2013exploring}. However, due to the complexity of the underwater environment and the optical characteristics of the water body (e.g., wavelength, distance-dependent attenuation and scattering), the underwater image will inevitably suffer from degradation (e.g., color deviation and low contrast~\cite{hu2022overview}). Therefore, how to restore a clear underwater image is particularly important for the development of the marine industry.

Traditional underwater image enhancement methods can be divided into physical model-free~\cite{zhang2022retinex,ancuti2017color,fu2014retinex,ancuti2012enhancing, Hitam2013Mixture} and physical model-based~\cite{li2016underwater, Galdran2015Automatic,wang2017single,peng2018generalization,drews2016underwater,qian2022real,liu2020investigating}. The physical model-free methods mainly adjust the pixel value (e.g., histogram equalization-based method~\cite{Hitam2013Mixture}) to improve the visual quality of the image. However, they ignore the underwater imaging mechanism, resulting in over-enhancement and over-saturation. The physical model-based methods are mainly based on various prior knowledge (e.g., underwater dark channel prior~\cite{drews2016underwater}) to estimate underwater imaging parameters (i.e., medium transmission and atmospheric light~\cite{anwar2020diving}), and then invert the physical model to obtain enhanced images. They also have limitations: 1) The estimated parameters are based on various prior conditions, but these prior conditions are not always accurate in different underwater environments (e.g., fuzzy prior~\cite{Peng2017Underwater} does not support clear underwater images). 2) It is a great challenge to estimate underwater imaging parameters accurately with physical methods.

In recent years, researchers try to use deep learning-based methods~\cite{perez2017deep,li2019underwater,li2020underwater,han2021deep,wang2021uiec, Mu2022StructureInferredBM} to enhance underwater images. Perez et al.~\cite{perez2017deep} form a pair of real-world underwater datasets for the first time, and use a simple CNN network to train the mapping relationship between degraded images and reference images. Han et al.~\cite{han2021deep} propose a deep supervised residual dense network. Wang et al.~\cite{wang2021uiec} propose UIEC $\hat{}$ 2-Net, which combines HSV and RGB color spaces, providing new ideas for future work. Although these methods are novel and exciting, their effects are not particularly attractive on the whole. There are two main reasons: 1) Most of them regard contrast enhancement and color restoration as the same, without special treatment of color separately. 2) Most of them neglect the underwater imaging mechanism and rely excessively on the feature learning ability of neural networks.

In order to remedy the above shortcomings, we propose an Adaptive Transmission and Dynamic Color guided network (named ATDCnet) for underwater image enhancement. By observing a large number of underwater datasets, we find that most underwater images are dominated by a single color. To deal with the color deviation problem, we design a Dynamic Color-guided Module (DCM) to post-process the enhanced image color. It can post-process the image color according to different water areas to restore the image color. Secondly, to exploit the knowledge of physics, we design an Adaptive Transmission-directed Module (ATM) to better guide the network. Further, we design an Encoder-Decoder-based Compensation (EDC) structure with attention and a multi-stage feature fusion mechanism to perform color restoration and contrast enhancement simultaneously. The main contributions can be summarized as follows:
\begin{itemize}
  \item We propose an Adaptive Transmission and Dynamic Color guided network (i.e., ATDCnet) applied for underwater image enhancement. This method focuses on color correction and contrast enhancement of images in different waters.
  \item To deal with the color deviation problem, we design a Dynamic Color-guided Module (DCM) to post-process the enhanced image color.
  \item To exploit the knowledge of physics, we design an Adaptive Transmission-directed Module (ATM) guiding the network to better Decoder.
  \item Extensive experiments on many benchmark datasets demonstrate that our ATDCnet has achieved state-of-the-art in terms of quantitative and visual performance.
\end{itemize}

\section{PROPOSED METHOD}
The underwater image degradation process can be represented by the modified Koschmieder light scanning model:
\begin{equation}
I_c(x) = J_c(x)T_c(x) + A_c(1-T_c(x)),
\end{equation}    
where $I_c(x)$ represents the observed image, $J_c(x)$ denotes the scene radiation, $x$ is the image pixel, $A_c$ defines the global background light, $c= \{r,g,b\}$ means the color channels. $T_c(x)=e^{-\beta_c d(x)}$ represents the transmission value, where $\beta_c$ is the channel-wise attenuation coefficient depending on water quality, $d(x)$ is the scene depth at pixel $x$.

In Fig.~\ref{fig:1}, we show the overall architecture of the proposed network. The network is mainly composed of an enhancement Encoder-Decoder-based Compensation (EDC) structure, Dynamic Color-guided Module (DCM), and Adaptive Transmission-directed Module (ATM). In the following content, we will briefly introduce the essential parts of our proposed network, mainly including the above three modules and related information fusion mechanisms.

\subsection{Adaptive Transmission-directed Module}

Thanks to the powerful feature extraction and representation capabilities of neural networks, we design an Adaptive Transmission-directed module (ATM) to better guide the network. We obtain an initial reverse medium transmission (RMT) map of raw underwater images via a robust general dark channel prior (DCP)~\cite{peng2018generalization}. Thus, it is a transmission-guided module which is not necessary to design the loss function of the ATM branch separately. The overall structure is shown in the ATM branch of Fig.~\ref{fig:1}. It is composed of three cascaded residual blocks, in which the number of channels is 64, 64, and 1 respectively. Using the network optimization method, more accurate transmission information can be obtained.

\subsection{Dynamic Color-guided Module}
\begin{figure}[t]
\centering
\includegraphics[width=\linewidth]{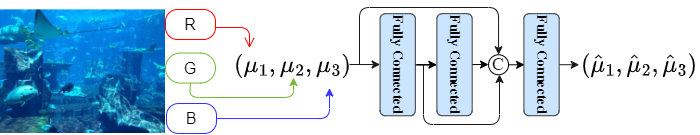}
\caption{Details of DCM. The input is the global average of the original underwater image according to the RGB channel, and the corrected attenuation coefficient is obtained after network optimization.}
\label{fig:2}
\end{figure}

We design a Dynamic Color-guided module (DCM) to post-process the enhanced image color. By observing a large number of underwater datasets, we find that most underwater images have a single color. In other words, a single color can reflect the general color information of the image. $(\mu_1,\mu_2,\mu_3)$ respectively represent the mean value of the red, green, and blue channels of the underwater image that are obtained by the following formulas:
\begin{equation}
    \begin{array}{l}
        \mu_1=\mathtt{mean}(I_r(x)),\ 
        \mu_2=\mathtt{mean}(I_g(x)),\ 
        \mu_3=\mathtt{mean}(I_b(x)),
    \end{array}
\end{equation}
% \begin{equation}
% \mu_1=\mathtt{mean}(I_r(x)), \mu_2=\mathtt{mean}(I_g(x)),\mu_3=\mathtt{mean}(I_b(x)),
% \end{equation}
where ``$\mathtt{mean}$" denotes the mean operator. Indeed, the global average of underwater image channels represents the overall color information. Therefore $(\mu_1,\mu_2,\mu_3)$ characterize the general color information of the observed underwater image. Then $(\mu_1,\mu_2,\mu_3)$ enter the DCM. After network optimization, three RGB channel color attenuation coefficients $(\hat{\mu_1},\hat{\mu_2},\hat{\mu_3})$ are obtained. Finally, the color is corrected by multiplying the color attenuation coefficient and the feature map. The overall structure of the DCM is shown in Fig.~\ref{fig:2}, which is composed of three fully connected layers. We concatenate the original input, the output of the first layer, and the output of the second layer according to the channel. This can effectively increase the available information on the network so that a more accurate color attenuation coefficient can be estimated. 

\subsection{Encoder-Decoder-based Compensation Structure}

In order to preserve the data fidelity and solve the problem of gradient disappearance, we take the residual block, which structure is shown in Fig.~\ref{fig:1}, as the basic component of the structure. In order to avoid unnecessary information loss, the convolution kernel and stride of all convolution layers are $3\times 3$ and 1 respectively. Such convolution operation will not change the image resolution. In addition, we also introduce the channel-attention (CA) mechanism, by assigning different weights to different channels to highlight more critical features. In the decoder stage, in order to make full use of the features of different stages ($F_1, F_2, F_3$ in Fig.~\ref{fig:1}), we design a multi-stage feature fusion mechanism. With the help of the attention mechanism, multi-stage features can increase more available information for the network, thus improving the network's performance. 

\subsection{Loss Function}
In order to achieve an effective balance between visual quality and quantitative scores, we adopt a linear combination of $\ell_2$ loss, perceptual loss, and SSIM loss. Specifically, $\ell_2$ loss measures the $\ell_2$ distance between the reconstructed image $J$ and the reference image $\hat{J}$ :
\begin{equation}
L_{\ell_2}=\frac{1}{N} \sum_{i=1}^{N} \lVert\hat{J_{i}}-J_{i}\rVert_{2},
\end{equation}    
where $J_{i}$ represents the pixel value at the reconstructed image position $i$, $\hat{J_{i}}$ represents the pixel value at the reference image position $i$. Since $\ell_2$ loss is difficult to capture high-level semantics, we introduce perceptual loss to evaluate the visual quality of images. It measures the $\ell_1$ distance between the reconstructed image $J$ and the reference image $\hat{J}$ in the feature space defined by VGG-19:
\begin{equation}
L_{perc}=\frac{1}{C_{k}H_{k}W_{k}} \sum_{i=1}^{N} \lVert \phi_{k}(J_{i})-\phi_{k}(\hat{J}_i) \rVert_1,
\end{equation} 
where $\phi_k$ represents the $k_{th}$ convolutional layer. N is the number of each batch in the training phase. $C_{k}, H_{k}, W_{k}$ represents the channel number, height, and width of the feature map at layer k of the VGG-19 network respectively. In our experiment, we calculate the perceptual loss at layer $relu5\_3$ of the VGG-19 network. 
In order to maintain the similarity of structure and texture between the reconstructed image $J$ and the reference image $\hat{J}$, we introduce SSIM loss: 
\begin{equation}
L_{SSIM}=1-\frac{1}{N} \sum_{i=1}^{N}SSIM(J_i, \hat{J}_i),
\end{equation} 
all losses act on the output stage of the network, and the total loss finally used for the training phase is expressed as follows:
\begin{equation}
L_{total}= \alpha L_{\ell_2}+\beta L_{perc}+\gamma L_{SSIM},
\end{equation} 
according to experience, we set $\alpha$, $\beta$, and $\gamma$ as 1, 0.01, and 100 respectively.

\begin{table}[t]
	\centering
	\caption{Ablation study of different settings, i.e., w/o CA and w/o $\hat{F}:=(\hat{F}_1,\hat{F}_3)$.}
	\begin{tabular}{c|c|c|c|c}
		\toprule
		Type & Model & PSNR$\uparrow$ & SSIM$\uparrow$ & UCIQE$\uparrow$ \\
		\hline
		A & \makecell[c]{w/o CA, w/o $\hat{F}$} &19.92 & 0.89 & 0.67 \\
		\hline
		B & \makecell[c]{w/o CA, with $\hat{F}$} &19.59 & 0.89 & 0.67 \\
		\hline
		C& \makecell[c]{with CA, w/o $\hat{F}$} &22.93 & 0.92 & 0.72 \\
		\hline
		D & \makecell[c]{Ours} &23.43 & 0.92 & 0.74 \\
		\bottomrule
	\end{tabular}
	\label{table:1}
\end{table}

\begin{figure}[t]
	\centering
	\includegraphics[width=1\linewidth]{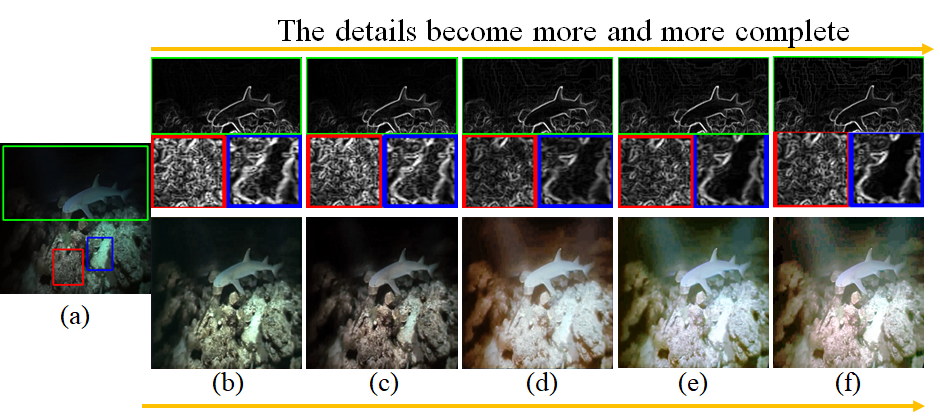}
	\caption{Ablation study of CA and $\hat{F}$ of EDC on UIEB dataset. (a) Input. (b) w/o CA + w/o $\hat{F}$. (c) w/o CA + with $\hat{F}$. (d) with CA + w/o $\hat{F}$. (e) ours. (f) Ground Truth. The first raw is the gradient map of the enhanced image, and the second raw is the enhanced image. The EDC with CA and $\hat{F}$ produces clearer and more complete image details.}
	\label{fig:3}
\end{figure}

\section{EXPERIMENTS}

\subsection{Setups}
\textbf{Datasets.} We evaluate the performance of the proposed method on two types of datasets: the first type has reference images (e.g., EUVP~\cite{islam2020fast}, UIEB~\cite{li2019underwater}, LSUI~\cite{peng2021u}, UFO-120~\cite{islam2020simultaneous}); The other is without reference image (e.g., C60~\cite{li2019underwater}, RUIE~\cite{liu2020real}, SQUID~\cite{berman2020underwater}). 

\textbf{Metrics.} We use three commonly used image evaluation metrics (i.e., Mean Square Error (MSE), Peak Signal to Noise Ratio (PSNR), and Structure Similarity Index (SSIM)) to compare different methods quantitatively. A higher PSNR or SSIM means that the enhanced image is closer to nature in terms of vision and structure. A lower MSE means that the image has a better reconstruction effect. In addition, we also introduce non-reference Underwater Image Quality Measure (UIQM)~\cite{panetta2015human} and Underwater Color Image Quality Evolution (UCIQE)~\cite{yang2015underwater}. A higher UIQM or UCIQE represents a better human visual perception.

\subsection{Ablation Study}
In this section, we will conduct ablation experiments in two kinds to study the role of each module. In the first kind, only the enhancement EDC structure is ablated ATM and DCM are fixed; In the second kind, ATM and DCM are ablated (EDC is fixed).

\textbf{Channel-attention (CA) + feature ($\hat{F}:=(\hat{F}_1,\hat{F}_3)$).} EDC is responsible for contrast enhancement and color restoration in our method. In order to improve the performance of this module, we introduce the CA and $\hat{F}:=(\hat{F}_1,\hat{F}_3)$. We conduct ablation experiments on EDC separately, and the experimental results are shown in Table~\ref{table:1}. Table~\ref{table:1} A and B show that without CA, $\hat{F}$ will damage the model's performance. Table~\ref{table:1} A and C show that the network performance has been greatly improved, which indicates that CA effectively helps the network select more critical features. Table~\ref{table:1} C and D illustrate that $\hat{F}$ will retain key features after CA screening. As can be seen, Fig.~\ref{fig:3}(e) with CA and $\hat{F}$ has more complete image details.

\begin{figure}[t]
\centering
\includegraphics[width=1\linewidth]{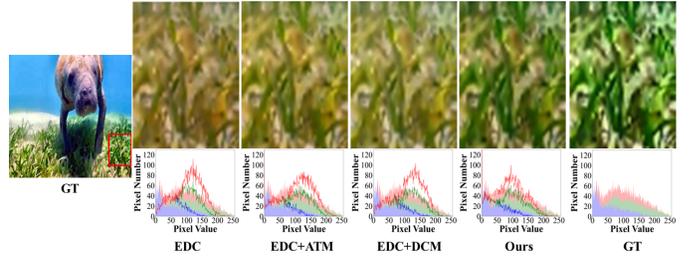}
\caption{Ablation study of the ATM and DCM on UIEB dataset. It can be seen from the pixel distribution map (The abscissa is the pixel value, and the ordinate is the number of pixels) that the complete model (Ours) with ATM and DCM is closer to Ground Truth. %This means that the enhanced image is more natural in contrast and color.
}
\label{fig:4}
\end{figure}

\textbf{EDC + ATM.} We combine the transmission information to better guide the EDC to enhance the image. Observing Table~\ref{table:2} (i) and (ii), PSNR and SSIM have improved significantly. However, the improvement of the visual effect is not obvious after observing Fig.~\ref{fig:4} (EDC) and Fig.~\ref{fig:4} (EDC+ATM). In low-level visual tasks, a high PSNR or SSIM does not necessarily represent a good image visual effect. After the introduction of the ATM, although the reconstruction effect has been improved (e.g. PSNR and SSIM have been improved), these reconstructed pixels do not necessarily conform to human visual perception.

\begin{table}[t]
\centering
\caption{Ablation study of the framework components on UIEB dataset} \label{tab:cap}
% \scalebox{0.9}{
\begin{tabular}{c c c c || c c c}
\toprule
Settings & EDC & ATM & DCM & PSNR$\uparrow$ & SSIM$\uparrow$ & UIQM$\uparrow$  \\
\hline
(i)  &  \Checkmark & \ & \ & 22.79 & 0.91 & 4.08  \\
(ii) &   \Checkmark & \Checkmark & \ & 23.29 & 0.92 & 4.10  \\
(iii) &   \Checkmark & \ & \Checkmark & 22.91 & 0.91 & 4.15  \\
(iv)  &  \Checkmark & \Checkmark & \Checkmark & 23.43 & 0.92 & 4.29  \\
\bottomrule
\end{tabular}
\label{table:2}
\end{table}

\begin{table*}[t]
\centering
\caption{Quantitative results in four different underwater datasets (i.e., EUVP, UIEB, UFO-120, LSUI).}
\begin{tabular}{c||c c c c c c c c c c}
 \toprule
  Datasets & Metrics & UDCP & Fusion & Water-Net & UGAN & FUnIE-GAN & Ucolor & USUIR & PUIE-Net & Ours \\
  \hline
         & PSNR$\uparrow$ & 16.38 & 17.61 & 20.14 & 23.51 & \underline{23.54} & 21.89 & 23.52 & 22.60 & \textbf{25.62} \\
   EUVP  & SSIM$\uparrow$ & 0.64 & 0.75 & 0.68 & \underline{0.81} & \underline{0.81} & 0.79 & \underline{0.81} & \underline{0.81} & \textbf{0.87} \\
         & MSE$\downarrow$  & 1990 & 1331 & 826 & 401 & \underline{398} & 505 & 399 & 438 & \textbf{239} \\ 
  \hline
         & PSNR$\uparrow$ & 13.05 & 17.60 & 19.11 & 20.17 & \underline{20.68} & 20.63 & 20.31 & 20.36 & \textbf{23.43} \\
   UIEB  & SSIM$\uparrow$ & 0.62 & 0.77 & 0.79 & 0.82 & 0.72 & 0.84 & 0.84 & \underline{0.88} & \textbf{0.92} \\
         & MSE$\downarrow$  & 3779 & 1331 & 1220 & 874 & 752 & 770 & 804 & \underline{730} & \textbf{416} \\ 
 \hline
         & PSNR$\uparrow$ & 18.26 & 14.58 & 22.46 & 23.45 & \underline{25.15} & 21.04 & 17.45 & 21.62 & \textbf{25.23} \\
UFO-120  & SSIM$\uparrow$ & 0.72 & 0.54 & 0.79 & \underline{0.80} & \textbf{0.82} & 0.66 & 0.69 & 0.74 & \textbf{0.82} \\
         & MSE$\downarrow$  & 1249 & 2968 & 458 & 393 & \underline{253} & 599 & 1327 & 514 & \textbf{232} \\ 
  \hline
         & PSNR$\uparrow$ & 12.66 & 14.48 & 17.73 & 19.79 & 19.37 & 21.56 & 18.64 & \underline{22.48} & \textbf{26.02} \\
   LSUI  & SSIM$\uparrow$ & 0.62 & 0.79 & 0.82 & 0.78 & \underline{0.84} & \underline{0.84} & 0.82 & \textbf{0.91} & \textbf{0.91} \\
         & MSE$\downarrow$  & 4529 & 3501 & 1361 & 883 & 946 & 563 & 1025 & \underline{493} & \textbf{238} \\ 
  \bottomrule
\end{tabular}
\label{table:3}
\end{table*}

\begin{table}[t]
\centering
\caption{Averaged unsupervised scores (i.e., UIQM, UCIQE) on real-world underwater datasets (i.e., C60, RUIE, SQUID) without reference
images.}
\setlength{\tabcolsep}{1pt}
\begin{tabular}{c || c | c | c |c |c | c }
\toprule
\multirow{2}*{Methods}& \multicolumn{2}{c|}{C60} &\multicolumn{2}{|c|}{RUIE} &\multicolumn{2}{|c}{SQUID} \\
\cline{2-7}
& UIQM $\uparrow$ & UCIQE $\uparrow$ & UIQM $\uparrow$ & UCIQE $\uparrow $& UIQM $\uparrow$ & UCIQE $\uparrow$ \\
\hline

UDCP &  5.55 & 0.65 & 5.34 & 0.60 & 4.96 & 0.59\\
Fusion & 5.79 & \textbf{0.76} & 5.41 & \textbf{0.76} & 5.35 & \textbf{0.79} \\
Water-Net & 5.63 & 0.66 & 4.69 & 0.62 & 5.25 & 0.47 \\
UGAN & 5.67 & 0.67 & 5.13 & 0.64 & 5.30 & 0.44 \\
FUnIE-GAN & 5.79 & 0.68 & 5.13 & 0.66 & 5.35 & 0.46 \\
Ucolor & \underline{5.80} & 0.67 & 5.40 & 0.63 & \textbf{5.79} & 0.51 \\
USUIR & 5.70 & 0.68 & \underline{5.48} & \underline{0.75} & 5.18 & \underline{0.66} \\
PUIE-Net & 5.46 & 0.62 & 4.94 & 0.59 & 5.56 & 0.52 \\
Ours & \textbf{5.83} & \underline{0.70} & \textbf{5.57} & 0.69 & \underline{5.69} & 0.55\\
\bottomrule
\end{tabular}
\label{table:4}
\end{table}

\begin{figure*}[t]
 \centering
 \begin{center}
     \begin{tabular}{c@{\extracolsep{0.1mm}}c@{\extracolsep{0.1mm}}c@{\extracolsep{0.1mm}}c@{\extracolsep{0.1mm}}c@{\extracolsep{0.1mm}}c@{\extracolsep{0.1mm}}c@{\extracolsep{0.1mm}}c}
         \rotatebox{90}{\scriptsize{~~~~~~EUVP}}
         \includegraphics[width=0.12\textwidth]{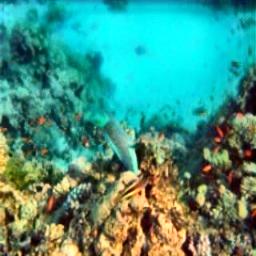}&
         \includegraphics[width=0.12\textwidth]{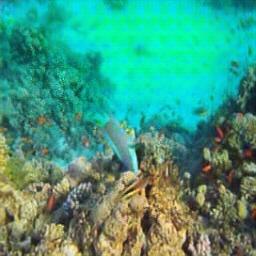}&
         \includegraphics[width=0.12\textwidth]{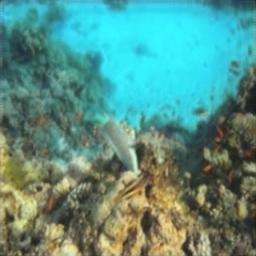}&
         \includegraphics[width=0.12\textwidth]{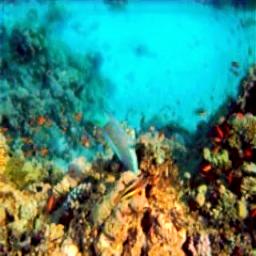}&
         \includegraphics[width=0.12\textwidth]{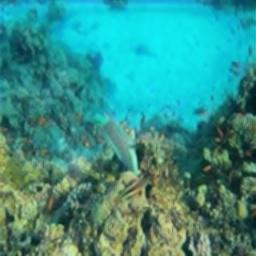}&
         \includegraphics[width=0.12\textwidth]{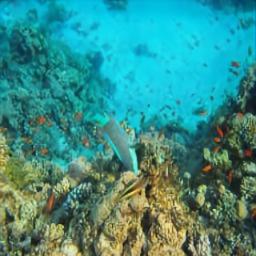}&
         \includegraphics[width=0.12\textwidth]{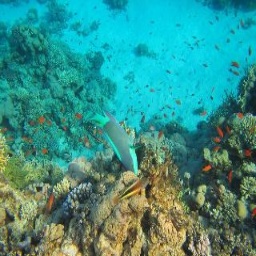}\\
         \rotatebox{90}{\scriptsize{~~~~~~~LSUI}}
         \includegraphics[width=0.12\textwidth]{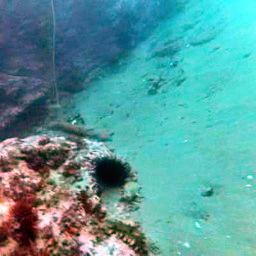}&
         \includegraphics[width=0.12\textwidth]{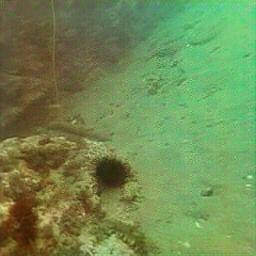}&
         \includegraphics[width=0.12\textwidth]{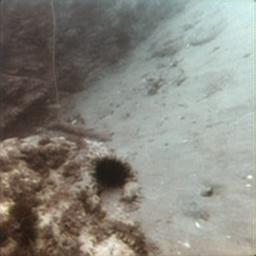}&
         \includegraphics[width=0.12\textwidth]{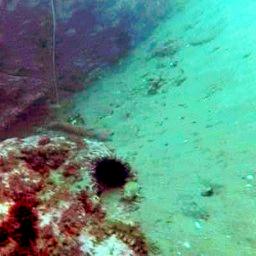}&
         \includegraphics[width=0.12\textwidth]{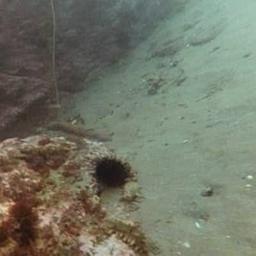}&
         \includegraphics[width=0.12\textwidth]{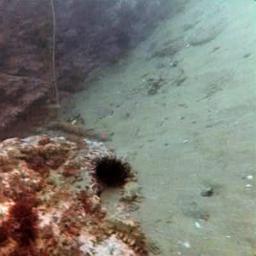}&
         \includegraphics[width=0.12\textwidth]{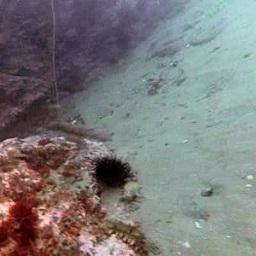}\\
         \rotatebox{90}{\scriptsize{~~~~UFO-120}}  
         \includegraphics[width=0.12\textwidth]{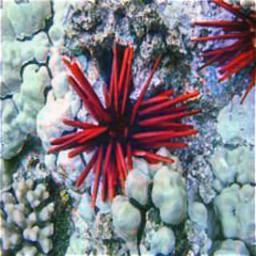}&
         \includegraphics[width=0.12\textwidth]{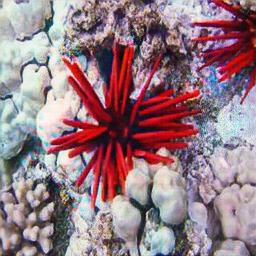}&
         \includegraphics[width=0.12\textwidth]{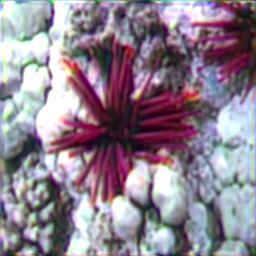}&
         \includegraphics[width=0.12\textwidth]{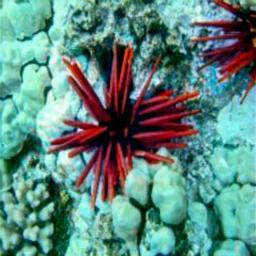}&
         \includegraphics[width=0.12\textwidth]{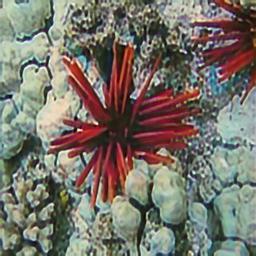}&
         \includegraphics[width=0.12\textwidth]{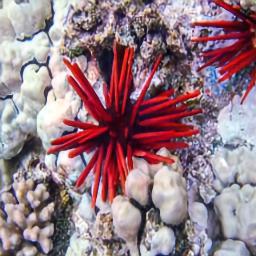}&
         \includegraphics[width=0.12\textwidth]{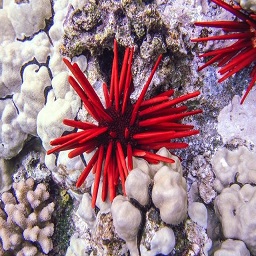}\\
         \rotatebox{90}{\scriptsize{~~~~UIEB}}  
         \includegraphics[width=0.12\textwidth]{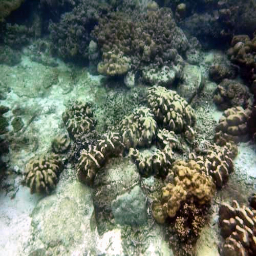}&
         \includegraphics[width=0.12\textwidth]{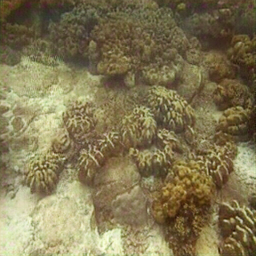}&
         \includegraphics[width=0.12\textwidth]{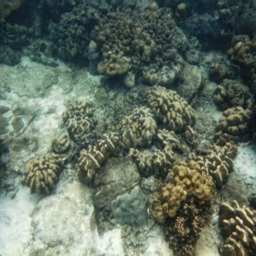}&
         \includegraphics[width=0.12\textwidth]{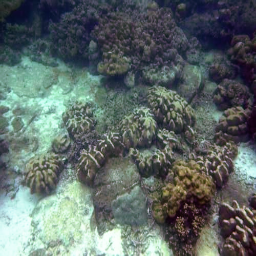}&
         \includegraphics[width=0.12\textwidth]{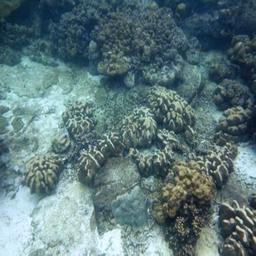}&
         \includegraphics[width=0.12\textwidth]{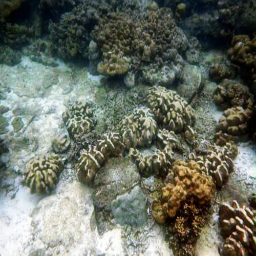}&
         \includegraphics[width=0.12\textwidth]{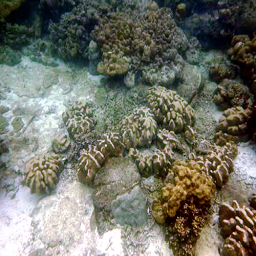}\\ 
         \footnotesize (a) Fusion & \footnotesize (b) FUnIE-GAN & \footnotesize (c) Ucolor  & \footnotesize (d) USUIR & \footnotesize (e) PUIE-Net & \footnotesize (f) Ours (ATDCnet) & \footnotesize (g)  Ground Truth
     \end{tabular}
 \end{center}
 \caption{Visual comparison with different underwater image enhancement methods on real-world datasets (i.e., EUVP, LSUI, UFO-120, UIEB). %These compared methods can not restore color or contrast very well. On the contrary, our method has an outstanding effect on color and contrast. %(a) Fusion (b) FUnIE-GAN  (c) Ucolor (d) USUIR (e) PUIE-Net (f) Ours (ATDCnet) (g) Ground Truth.
 }
 \label{fig:5}
 \end{figure*}

\textbf{EDC + DCM.} We conduct post-processing on the enhanced image color. In Table~\ref{table:2}, by comparing Table~\ref{table:2} (ii) and (iii), we can see that UIQM has been effectively improved. Comparing Fig.~\ref{fig:4} (EDC+ATM) and Fig.~\ref{fig:4} (EDC+DCM), it can be seen that DCM has the ability to restore image color.

\textbf{EDC + ATM + DCM.} According to Table~\ref{table:2} (iv), after combining the three modules, various metrics have been significantly improved. We believe that this is the result of the assistance between the three modules. After the introduction of the ATM, the image reconstruction effect is improved, but these reconstructed pixels do not conform to human visual perception. When the DCM is added, it can assist the ATM to reconstruct pixels better. The reconstructed pixel has both a better visual and reconstruction effect, so the final image will be more natural. Looking at Fig.~\ref{fig:4} (Ours) and Fig.~\ref{fig:4} (GT), the enhanced image is very close to the reference image.

\subsection{Comparison Results}
We make quantitative and qualitative comparisons with eight state-of-the-art methods, including traditional methods 
(e.g., UDCP~\cite{drews2016underwater}, 
Fusion~\cite{ancuti2012enhancing}), CNN-based methods (e.g., Water-Net~\cite{li2019underwater}, Ucolor~\cite{li2021underwater}, USUIR~\cite{fu2022unsupervised}, PUIE-Net~\cite{fu2022uncertainty}), and GAN-based methods (e.g., UGAN~\cite{fabbri2018enhancing}, FUnIE-GAN~\cite{islam2020fast}) on seven datasets.

\textbf{Quantitative comparison.} We test the supervised metrics in four underwater datasets (i.e., EUVP, UIEB, LSUI, UFO-120). The average scores of MSE, PSNR, and SSIM are presented in Table~\ref{table:3}. As shown in Table~\ref{table:3}, in each line, we use black bold to indicate the best and underline to indicate the second. First of all, it is obvious that the average score of our design method is much higher than other methods. Secondly, the robustness of these state-of-the-art methods is uneven (e.g., FUnIE-GAN performs poorly on LSUI but better on the other three datasets). On the contrary, our method can be well generalized on all datasets.

In addition, we also conduct experiments on challenging datasets (i.e., C60, RUIE, SQUID). The results are presented in Table~\ref{table:4}. Looking at Table~\ref{table:4}, our method obtains two first (UIQM: C60 and RUIE) and two seconds (UCIQE: C60. UIQM: SQUID). In general, our method has better generalization performance than other methods. In addition, We find that although the traditional methods (e.g., Fusion) have lower supervised metrics (e.g., PSNR), they have higher unsupervised metrics (e.g., UCIQE).

\textbf{Qualitative Comparisons.} In Fig.~\ref{fig:5}, we provide the enhanced results of the designed method and the method with relatively high PSNR and SSIM scores (i.e., Fusion~\cite{ancuti2012enhancing}, FUnIE-GAN~\cite{islam2020fast}, Ucolor~\cite{li2021underwater}, PUIE-Net~\cite{fu2022uncertainty}, USUIR~\cite{fu2022unsupervised}) in Table~\ref{table:3}. It can be seen from Fig.~\ref{fig:5} that the traditional method (e.g., Fusion~\cite{ancuti2012enhancing}) has over-enhanced the image; The image contrast enhanced by the GAN-based method (e.g., FUnIE-GAN~\cite{islam2020fast}) is obviously insufficient. The CNN-based methods (e.g., USUIR~\cite{fu2022unsupervised}, PUIE-Net~\cite{fu2022uncertainty}) have significantly improved the contrast, but the image color has not been effectively restored. On the contrary, our enhanced image is more attractive in terms of color and contrast. Due to space limitations, we provide more qualitative analysis in the supplementary materials. 

\section{CONCLUSION}

We propose a new underwater image enhancement model. On the basis of improving image contrast by the Encoder-Decoder-based Compensation (EDC) structure, the deep processing of color is realized by the Dynamic Color-guided Module (DCM). In addition, domain knowledge is incorporated into the network through the Adaptive Transmission-directed Module (ATM). Extensive experiments on different benchmark datasets have proved the superiority of our solution. The ablation study verified the effectiveness of the key components of our method.

%\section*{Acknowledgment}
%This work is supported by Natural Science Foundation of China (Grant No. U20A20196, 62202429) and Zhejiang Provincial Natural Science Foundation of China under Grant No. LR21F020002, LY23F020024.

\footnotesize
\bibliographystyle{IEEEbib}
\bibliography{reference}
\end{document}